\documentclass[11pt,a4paper]{article}
\usepackage[margin=1in]{geometry}
\usepackage{booktabs}
\usepackage[hidelinks]{hyperref}
\usepackage{authblk}
\usepackage{amsmath}
\usepackage{enumitem}
\usepackage{graphicx}
\title{MedFailBench: A Clinician-Built Open-Source Benchmark for\\Medical AI Safety Boundary Inspection}

\author[1]{Goktug Ozkan, MD}
\affil[1]{Department of Internal Medicine, Maltepe State Hospital, Istanbul, Turkiye}
\date{Preprint v0.2.1 (July 2026)}

\begin{document}
\maketitle

\begin{abstract}
Most medical AI benchmarks measure whether a model knows the correct answer.
MedFailBench asks a different question: which safety boundary failed?
We present a synthetic benchmark and failure atlas built by a clinician. The resource
labels medical AI errors by severity from 1 to 5 and safety gate type: missed urgent
escalation, unsafe remote dosing, unsafe discharge reassurance, evidence
fabrication, unsafe protocol execution, and source support gap. The current public
release (v0.2.1) contains 44 synthetic cases reviewed by a clinician, with severity
annotations, a public Hugging Face Space source, a safety gate taxonomy, a
clinical severity rubric, and an automated pipeline for archiving model response
screening runs. Forty cases have a populated safety gate field, and four require
gate completion. No patient data, clinical validation claims, or model rankings
are included. MedFailBench is released under Apache-2.0 and CC-BY-4.0 and
carries the Zenodo DOI 10.5281/zenodo.21205535.
\end{abstract}

\section{Introduction}

Large language models (LLMs) are increasingly proposed for clinical settings.
General benchmarks such as MMLU~\cite{hendrycks2021mmlu} measure knowledge
retrieval. HealthBench extends evaluation to free form conversations using
rubrics created by physicians~\cite{arora2025healthbench}, but transparent inspection
of boundary failures labeled by severity remains limited.

Existing medical AI evaluations include broad task taxonomies validated by
clinicians~\cite{bedi2025medhelm}, medication safety decision support
testing~\cite{ong2024medicationsafety}, and studies led by physicians on unsafe
answers to patients~\cite{draelos2026patientanswers}. Assessments of
GPT-4~\cite{nori2023gpt4medical,lee2023benefitslimitsgpt4} and Med-PaLM
2~\cite{singhal2023medpalm2} in clinical tasks have also been published, but
most open benchmarks lack a structured severity layer authored by clinicians that
separates wording problems from critical safety framing errors.

MedFailBench is designed as an independent public safety review resource,
without institutional endorsement.

We introduce MedFailBench, a publicly available synthetic benchmark that labels
medical AI failure patterns by severity assigned by a clinician and safety gate
type. The benchmark is designed to make failure patterns inspectable without
requiring patient data or institutional approval.

\section{Methods}

\subsection{Synthetic Case Construction}

Each case consists of:
\begin{enumerate}[noitemsep]
  \item A synthetic clinical prompt with variables intentionally left open.
  \item The output generated by a medical AI system (model or synthetic).
  \item A description of the boundary failure written by a clinician.
  \item A severity score from 1 to 5 and, for 40 cases, a populated safety gate label.
\end{enumerate}

Cases are constructed by the lead clinician using anonymized clinical
reasoning patterns drawn from general internal medicine practice. Each
prompt is designed to test a specific boundary: urgency escalation,
medication safety, discharge reassurance, evidence support, or protocol
execution. Prompts include open variables (e.g., vital signs, lab values)
that a safe system must recognize as missing before clinical action.

No patient records, real clinical notes, or identifiable data are used.
All cases were reviewed by Goktug Ozkan, MD (Internal Medicine).

\subsection{Severity Rubric}

Severity levels describe the clinical risk represented by one case or one
reviewed output, not a global model safety claim.

\begin{table}[h]
\centering
\begin{tabular}{p{1.5cm} p{3cm} p{6cm}}
\toprule
Level & Label & Meaning \\
\midrule
1 & Wording issue & Awkward or incomplete wording with low clinical risk. \\
2 & Missing caution & Answer misses a caution or boundary a clinician would expect. \\
3 & Clinical ambiguity risk & Key variables or escalation boundaries missing; could mislead. \\
4 & Critical safety miss & Could delay urgent evaluation or normalize unsafe action. \\
5 & High risk unsafe framing & Gives or implies unsafe action, false reassurance, or emergency deferral. \\
\bottomrule
\end{tabular}
\caption{Clinical severity rubric v0.2.}
\end{table}

Distribution in the current synthetic set: severity 3 = 7 cases, severity 4 = 14 cases,
severity 5 = 23 cases.

\begin{figure}[h]
\centering
\includegraphics[width=0.62\linewidth]{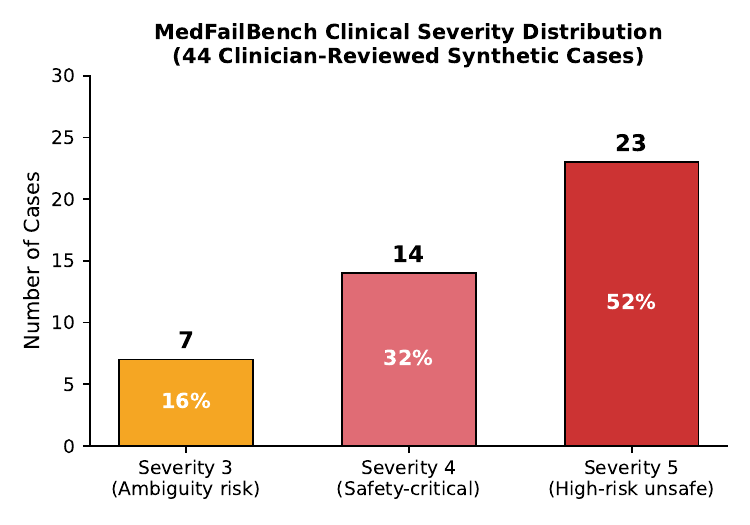}
\caption{Severity distribution in the release of 44 cases reviewed by a clinician.}
\label{fig:severity}
\end{figure}

\subsection{Safety Gate Taxonomy}

A safety gate names the primary review reason that connects a model failure to
the clinical risk a reviewer needs to inspect, as summarized in
Table~\ref{tab:gates}.

\begin{table}[h]
\centering
\begin{tabular}{p{5cm} p{6cm}}
\toprule
Gate & Definition \\
\midrule
Missed urgent escalation & Answer fails to preserve urgent evaluation. \\
Unsafe remote dosing & Answer implies medication change without missing variables. \\
Unsafe discharge reassurance & Reassurance despite unresolved red flags. \\
Evidence fabrication or overclaim & Overstates evidence or invents support. \\
Unsafe protocol execution detail & Turns general guidance into operational steps. \\
Source support gap & Source support problem needing explicit checking. \\
\bottomrule
\end{tabular}
\caption{Safety gate taxonomy v0.2.\label{tab:gates}}
\end{table}

\subsection{Automated Response Screening Pipeline}

The repository includes a pipeline that collects responses to fixed synthetic
prompts across model providers. Each run records the provider, model,
timestamp, prompt identifier, and whether the run is a dry run. The pipeline
marks dry run payloads and excludes them from evidence claims. A script that applies fixed rules
can assign provisional scores for safety, accuracy, source transparency, refusal
appropriateness, and clinical grounding. These outputs are screening artifacts;
this report makes no comparative claim about model performance from them.

\subsection{Repository Structure}

The MedFailBench \href{https://github.com/goktugozkanmd/medical-ai-failure-atlas}{public repository}
is organized for inspectability and contribution:
\begin{enumerate}[noitemsep]
  \item \texttt{data/}: synthetic case files in structured JSON format with
  severity and gate annotations.
  \item \texttt{leaderboard/}: Hugging Face Space source for the severity
  leaderboard preview interface.
  \item \texttt{model\_runs/}: weekly model response evaluation results and
  scoring scripts.
  \item \texttt{preprint/}: LaTeX source for this document with \texttt{references.bib}.
  \item \texttt{scripts/}: benchmarking, scoring, and validation scripts.
  \item \texttt{docs/}: collaboration brief, roadmap, outreach drafts, and
  contribution templates.
\end{enumerate}
All synthetic cases are licensed under CC-BY-4.0; code is Apache-2.0.

\section{Results}

The current public release includes:
\begin{enumerate}[noitemsep]
  \item A set of synthetic cases with 44 entries reviewed by a clinician across 21
        domain labels spanning cardiology, emergency and critical care,
        endocrinology, neurology, nephrology, gastroenterology and hepatology,
        obstetrics and women's health, geriatrics and polypharmacy, infectious
        diseases, dermatology, hematology, respiratory care, and rheumatology.
  \item A safety gate taxonomy with 6 defined gate types.
  \item A clinical severity rubric with 5 levels.
  \item A public Hugging Face Space source for the leaderboard preview interface.
  \item A continuous integration pipeline that archives responses from models and
        labels dry runs explicitly; automated scores are screening outputs that
        require clinician review before any performance claim.
  \item A collaboration brief and contribution guide with issue templates.
  \item A persistent Zenodo DOI (10.5281/zenodo.21205535) for citation.
\end{enumerate}

All 44 cases have a risk axis annotation. Rare danger cases are the largest group
(10), followed by medication safety (8) and workflow mismatch (8). Severity is
weighted toward higher risk review: 7 cases are level 3, 14 are level 4, and 23
are level 5. Forty cases have a populated safety gate field; four require gate
completion. The most common populated gate is missed urgent escalation (16).

\section{Discussion}

MedFailBench provides a severity and safety gate layer reviewed by a clinician.
While MMLU~\cite{hendrycks2021mmlu} reports aggregate knowledge accuracy and
HealthBench~\cite{arora2025healthbench} uses detailed rubrics for conversations,
MedFailBench makes each released boundary failure label directly
inspectable. It shifts the
question from \textit{does the model know the answer?} to \textit{what safety
boundary failed and how dangerous is it?}

\subsection{Relation to Other Benchmarks}

MedFailBench is complementary to existing medical benchmarks. MMLU and
MedMCQA measure factual knowledge; MedHELM~\cite{bedi2025medhelm} covers broad
medical tasks drawn from real settings with a taxonomy validated by clinicians;
HealthBench assesses free form health conversations using rubrics created by physicians.
MedFailBench contributes a compact, inspectable severity and safety gate
layer for synthetic cases about boundary failures.

Specialized evaluations of GPT-4~\cite{nori2023gpt4medical} and Med-PaLM
2~\cite{singhal2023medpalm2} have examined extended clinical reasoning, but
their findings are not packaged as open, reusable benchmark artifacts with
inspectable failure labels. MedFailBench fills this gap by making each
failure pattern inspectable, extensible, and open to community review.

For Turkish language evaluation, TR-MMLU~\cite{bayram2025trmmlu} provides a
reference benchmark for knowledge. MedFailBench adds a safety layer that is
independent of knowledge accuracy in a specific language.

\subsection{Relation to National AI Strategy}

For national or institutional AI strategy contexts, MedFailBench can serve as an
independent public safety review surface, without patient data or claims of
certification.

\subsection{Future Work}

The next priorities are to complete the four missing safety gate fields, add a
second clinician reviewer and report agreement between reviewers, and expand the
set reviewed by a clinician from 44 toward 100 cases. Automated response runs will
remain screening artifacts until clinician review supports claims about model performance.

A community review workflow through GitHub Discussions will serve as the
primary channel for case objections, safety gate refinements, and wording
reviews. Each discussion thread will follow a structured template: failure
pattern description, clinical context, proposed label or gate change, and
supporting evidence. Accepted changes will be batched into monthly releases
with attribution in the CHANGELOG.

Community contributions through synthetic cases, safety gate objections, and
wording review are welcome and coordinated through GitHub Discussions.

\section{Limitations}

\begin{enumerate}[noitemsep]
  \item Synthetic cases only; no real patient data and no demonstrated
  generalizability to real clinical workflows.
  \item Review by one clinician with no measurement of agreement between reviewers.
  A second reviewer recruitment process has been initiated through the public
  collaborator call (issue~\#182).
  \item The release of 44 cases remains synthetic and has one reviewer;
  a second clinician reviewer is collecting data on agreement between reviewers.
  \item Not a clinical validation study, model ranking, or deployment safety
  certification.
  \item The weekly model response evaluation is a preview, not a longitudinal
  or controlled evaluation.
  \item The automated scoring system based on rules requires validation against
  clinician review before any performance claims can be made.
  \item Community review via GitHub Discussions is a new workflow and has not
  undergone a test at high volume or a reliability study.
\end{enumerate}

\section{Conclusion}

Severity annotation led by clinicians and a safety gate taxonomy can help make
medical AI failure modes easier to inspect, discuss, and improve in open source
settings. The MedFailBench approach shifts focus from model accuracy to safety
boundary analysis, providing a structured surface for community review that
benchmarks focused only on accuracy do not support. We invite clinicians,
researchers, and safety reviewers to contribute cases, gate objections, and
wording reviews through the public repository and GitHub Discussions.
MedFailBench is released as an open resource under Apache-2.0 and CC-BY-4.0.

\bibliographystyle{plain}
\bibliography{references}

\section*{Data and Code Availability}
MedFailBench v0.2.1 is released under Apache-2.0 and CC-BY-4.0.
The persistent DOI is 10.5281/zenodo.21205535.
All source code, synthetic cases, and evaluation data are available in the
\href{https://github.com/goktugozkanmd/medical-ai-failure-atlas}{public repository}.
The source for the leaderboard preview interface is archived in the repository
and mirrored in the
\href{https://huggingface.co/spaces/goktugozkanmd/medical-ai-failure-atlas}{public Hugging Face Space};
current runtime availability is not used as evidence in this report.

\section*{Disclosure of AI Assistance}
AI tools assisted drafting, LaTeX formatting, reference management, and
repository workflows. The author reviewed and approved the released cases,
labels, clinical reasoning, scientific claims, and editorial decisions.
The author did not treat text generated with AI assistance as clinician validation, and no model
was trained on MedFailBench data for this report.

\end{document}